\documentclass[conference]{IEEEtran}
\IEEEoverridecommandlockouts
\usepackage{cite}
\usepackage{amsmath,amssymb,amsfonts}
\usepackage{algorithmic}
\usepackage{graphicx}
\usepackage{textcomp}
\usepackage{xcolor}

\usepackage{color}
\usepackage{makecell}

\usepackage{times}
\usepackage{soul}
\usepackage{url}
\usepackage[utf8]{inputenc}
\usepackage[small]{caption}
\usepackage{graphicx}
\usepackage{amsmath}
\usepackage{amsthm}
\usepackage{booktabs}
\usepackage{algorithm}
\usepackage{algorithmic}
\usepackage{makecell}
\usepackage{multirow}
\usepackage{color}
\usepackage{amssymb}
\usepackage{array}
\usepackage{booktabs}
\usepackage{colortbl}
\usepackage{tabularx}

\usepackage{algorithm}
\usepackage{algorithmic}
\usepackage{amsthm,amsmath,amssymb}
\usepackage{mathrsfs}
\usepackage{xcolor}
\usepackage{utfsym}
\usepackage{multirow}
\usepackage{makecell}

\def\BibTeX{{\rm B\kern-.05em{\sc i\kern-.025em b}\kern-.08em
    T\kern-.1667em\lower.7ex\hbox{E}\kern-.125emX}}

\makeatletter
\newcommand{\linebreakand}{%
  \end{@IEEEauthorhalign}
  \hfill\mbox{}\par
  \mbox{}\hfill\begin{@IEEEauthorhalign}
}
\makeatother

\begin{document}

\title{A Simple and Better Baseline for Visual Grounding}

\author{
\IEEEauthorblockN{Jingchao Wang}
\IEEEauthorblockA{\textit{School of Data Science and Engineering} \\
\textit{East China Normal University}\\
Shanghai, China \\
jcwang@stu.ecnu.edu.cn}
\and

\IEEEauthorblockN{Wenlong Zhang}
\IEEEauthorblockA{\textit{OpenScience Lab} \\
\textit{Shanghai AI Laboratory}\\
Shanghai, China \\
zhangwenlong@pjlab.org.cn}

\and
\IEEEauthorblockN{Dingjiang Huang$^{\ast}$}
\IEEEauthorblockA{\textit{School of Data Science and Engineering} \\
\textit{East China Normal University}\\
Shanghai, China \\
djhuang@dase.ecnu.edu.cn}

\linebreakand
\IEEEauthorblockN{Hong Wang$^{\ast}$}
\IEEEauthorblockA{\textit{School of Life Science and Technology} \\
\textit{Xi'an Jiaotong University} \\
Xi'an, China\\
hongwang01@xjtu.edu.cn}
\and
\IEEEauthorblockN{Yefeng Zheng}
\IEEEauthorblockA{\textit{Medical Artificial Intelligence Laboratory} \\
\textit{Westlake University}\\
 Hangzhou, China \\
zhengyefeng@westlake.edu.cn}
\thanks{${\ast}$ Corresponding Author.}
\thanks{This work was partially supported by the National Natural Science Foundation of China under Grant 62072185, U1711262, and Young Elite Scientists Sponsorship Program by CAST 2023QNRC001.}
\thanks{Published in ICME2025.}
}

\maketitle

\begin{abstract}

Visual grounding aims to predict the locations of target objects specified by textual descriptions. For this task with linguistic and visual modalities, there is a latest research line that focuses on only selecting the linguistic-relevant visual regions for object localization to reduce the computational overhead. Albeit achieving impressive performance, it is iteratively performed on different image scales, and at every iteration, linguistic features and visual features need to be stored in a cache, incurring extra overhead. To facilitate the implementation, in this paper, we propose a feature selection-based simple yet effective baseline for visual grounding, called FSVG. Specifically, we directly encapsulate the linguistic and visual modalities into an overall network architecture without complicated iterative procedures, and utilize the language in parallel as guidance to facilitate the interaction between linguistic modal and visual modal for extracting effective visual features. Furthermore, to reduce the computational cost, during the visual feature learning, we introduce a similarity-based feature selection mechanism to only exploit language-related visual features for faster prediction. Extensive experiments conducted on several benchmark datasets comprehensively substantiate that the proposed FSVG achieves a better balance between accuracy and efficiency beyond the current state-of-the-art methods. Code is available at \url{https://github.com/jcwang0602/FSVG}  .

\end{abstract}

\section{Introduction}
Visual grounding, also known as referring expression comprehension or phrase grounding, is a fundamental procedure in the field of vision-language integration, and it plays a great role in visual question answering and visual language navigation tasks~\cite{scanformer,cyco,selfeq,pvd,zeroshot_rec}. For visual grounding, the goal is to localize target objects or regions within an image specified by natural language descriptions.

Driven by the exciting success of Transformer in the field of computer vision and natural language processing~\cite{transformer}, it has been widely adopted in this visual grounding task. Currently, one mainstream research paradigm is sequentially composed of two core procedures, including using pretrained visual backbone networks and linguistic backbone networks to extract features for image and text modalities, respectively, and exploiting the Transformer encoder to achieve cross-modal feature fusion~\cite{transvg++} (see Fig.~\ref{fig: pipelin} (b)). Albeit obtaining promising performance, this research line generally suffers from a limitation that due to the insufficient interaction between two modalities during the first feature extraction procedure, the extracted visual features for the subsequent localization prediction may not align with the semantics of the natural language expression very well~\cite{qrnet,lgrnet}. To alleviate this issue, ~\cite{qrnet} proposed a guidance-based query-modulated refinement network QRNet to dynamically compute query-dependent visual attention in order to promote the extraction of meaningful visual features and make them consistent with text semantics (see Fig.~\ref{fig: pipelin}(c)). Nevertheless, it contains a complicated cross-modal fusion process, which leads to a certain computational cost.
Besides, most of these existing methods extract visual features by traversing images for localization. Actually, the images generally contain redundant information that is not relevant to target objects designated by textual descriptions. Such dense perception manner inevitably brings additional computational overhead.

Very recently, instead of adopting the dense perception of images, \cite{scanformer} proposed to eliminate linguistic-irrelevant redundant visual regions to further improve the model efficiency. In this work, the authors constructed a coarse-to-fine image perception framework, ScanFormer, that iteratively localizes target objects at different image scales. At every iteration, linguistic features and visual features are stored as the cache to guide the selection of linguistic-relevant visual patches. Although ScanFormer indeed strikes a better balance between localization accuracy and model efficiency beyond the existing methods for visual grounding, it relies on multiple iterations and the cache mechanism for multiple predictions at different scales, which is unfriendly for implementation.

\begin{figure}[t]
    \centering
    \includegraphics[width=0.8\linewidth]{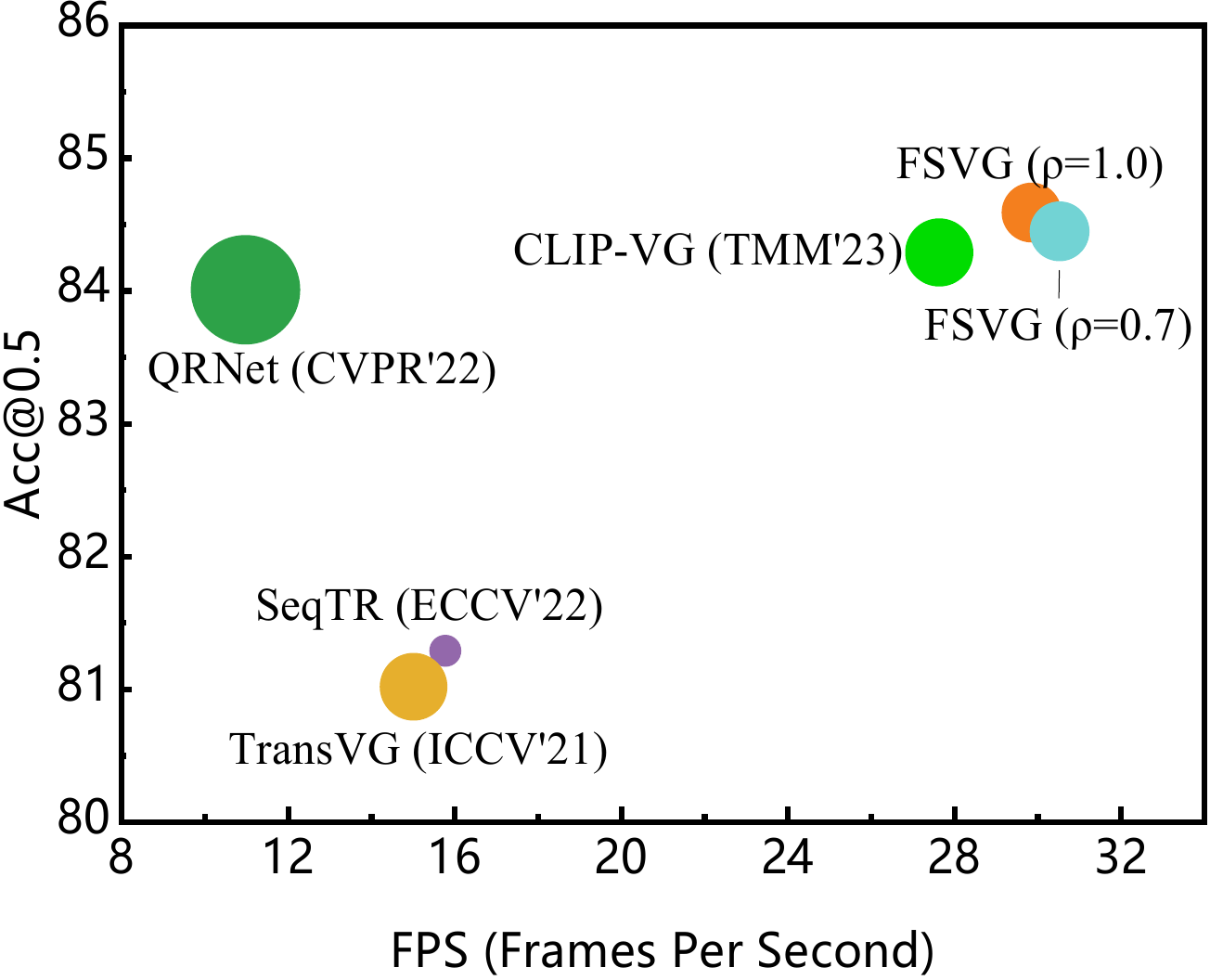}
    \vspace{-1mm}
    \caption{Comparison of accuracy and efficiency on the widely-adopted RefCOCO val set. The circle size is proportional to the number of model parameters. As seen, our FSVG strikes a better balance between performance and inference speed with comparable model parameters. $\rho$ denotes the ratio of visual feature selection. The lower the value, the fewer visual features are selected for faster prediction.}
    \vspace{-3mm}
    \label{fig: fps}
\end{figure}

Against these aforementioned issues, inspired by the selective perception of images paradigm adopted in ScanFormer, in this paper, we directly start from the feature level and aim to develop a simpler framework without complicated iterative procedures to {directly recognize the visual features that are weakly related or unrelated to natural language expressions and then more efficiently achieve the prediction by discarding these unimportant visual features.}
Specifically, instead of adopting the serial pipeline, \emph{i.e.}, multi-modal feature extraction first and then cross-modal feature fusion, we construct a parallel structure that directly feeds both visual tokens and language tokens to an overall network architecture (see Fig.~\ref{fig: pipelin}(d)). This seemingly simple and intuitive manner has two potential merits: 
1) linguistic features would be propagated through the whole visual feature extraction process; 
2) it provides the opportunity for multi-modal information interaction at the early stage of visual feature extraction. With the blessing of such double advantages, the visual feature learning would proceed in a right direction that aligns with the textual semantic information, and there is no need to additionally design the cross-modal fusion module after feature extraction like the existing serial processing paradigm. Furthermore, to accelerate computing, we incorporate a feature selection mechanism, which utilizes the similarity between visual features and linguistic features to help select linguistic-relevant visual features and discard the useless representation for faster prediction. 
Our main contributions are three-fold:
\begin{figure}[t]
    \centering
    \includegraphics[width=\linewidth]{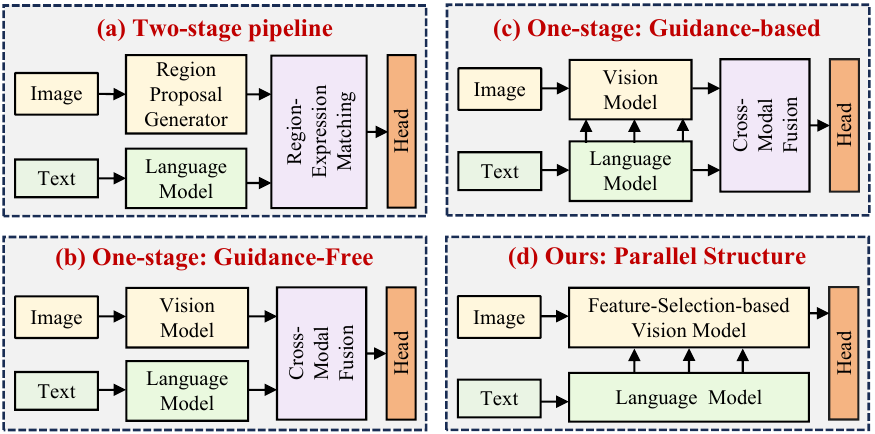}
    \caption{Comparisons of different pipelines for visual grounding.}
    \vspace{-3mm}
    \label{fig: pipelin}
    \vspace{-2mm}
\end{figure}

\begin{itemize}

\item For the visual grounding task, we specifically propose a simple and parallel structure to make linguistic semantics fully propagate through the entire visual feature extraction process, which would guide the effective extraction of visual features and enforce them to align with the textual semantics.

\item To further reduce the computational cost, we design a feature selection mechanism to capture linguistic-relevant visual features for faster localization prediction.

\item  Based on four mainstream datasets, our FSVG accomplishes a better balance between accuracy and efficiency with comparable model parameters as presented in Fig.~\ref{fig: fps}.

\end{itemize}

\begin{figure*}[!h]
    \centering
    \includegraphics[width=0.9\linewidth]{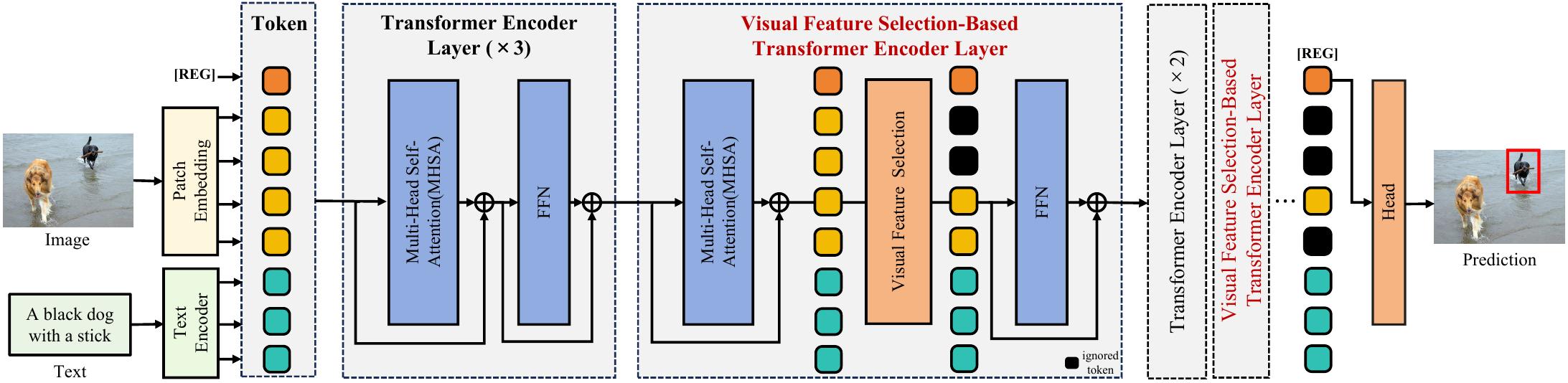}
    \caption{The entire architecture of the proposed FSVG which directly takes the concatenation of visual tokens and linguistic tokens as the input and consists of alternating vanilla Transformer layers and visual feature selection-based Transformer layers for faster 
 localization prediction.}
    \label{fig: fsvg}
     \vspace{-2mm}
\end{figure*}

\section{Methodology}

In this section, we construct the entire feature selection-based visual grounding framework, called FSVG. As presented in Fig.~\ref{fig: fsvg}, it mainly consists of three parts: 1) feature selection-based visual backbone network for linguistic-relevant visual feature extraction with the input as the concatenation of visual token $T_{v}$, learnable embedding (\emph{i.e.}, [REG] token), and linguistic token $T_{l}$, 2) language backbone network for textual feature extraction, and 3) the head structure with the [REG] token as the input for predicting the bounding box of target objects specified by the natural language expression. The details are described as follows.

\subsection{Parallel Multi-Modal Interaction for Visual Learning}
To guide the effective extraction of visual features that correspond to the natural language expression, we abandon the existing serial paradigm that is sequentially composed of multi-modal feature extraction and cross-modal feature fusion, and propose a parallel structure that makes the linguistic features propagate through the entire visual feature extraction process for providing comprehensive guidance. 

Given an input RGB image $X\in\mathbb{R}^{H\times W \times3}$ and the corresponding textual description, we first adopt the patch embedding layer and text encoder of CLIP to tokenize them as ${T_v \in \mathbb{R}^{N_v \times D}}$ and ${T_l \in \mathbb{R}^{N_l \times D}}$, respectively. 
Here $W$ and $H$ are the width and height of the image; 
$N_{v}=HW/P^2$ is the number of vision tokens; 
$P$ is the patch size;
 $N_l$ is the number of language tokens; 
 $D$ is the embedding dimension of every token.
Then, we concatenate the [REG] token $T_{REG} \in \mathbb{R}^{1 \times D}$ (a learnable embedding), visual tokens ${T_v \in \mathbb{R}^{N_v \times D}}$, and language tokens ${T_v \in \mathbb{R}^{N_l \times D}}$ in the first dimension and encapsulate it as the input:
\begin{equation}
\label{eq1}
    {T_{rvl}}=\mathrm{Concat}[T_{REG},T_{v},T_{l}],
\end{equation}
where ${T_{rvl} \in \mathbb{R}^{(1+ N_v + N_l) \times D}}$ is the input token sequence.

Inspired by the powerful relationship modeling capability of the self-attention mechanism involved in Transformer, it is natural to utilize this structure to achieve the multi-modal information interaction between text features and vision features. Specifically, as shown in Fig.~\ref{fig: fsvg}, we feed the encapsulated token sequence $T_{rvl}$ into a widely-adopted CLIP-ViT backbone. However, different from the vanilla ViT which stacked several blocks with the same Transformer encoder layers, in our FSVG, after every three Transformer blocks, we introduce a feature selection (FS) mechanism to only select linguistic-relevant visual features fed to the next block for the higher computational efficiency, 

Specifically, for the Transformer encoder layer without FS, it contains two computation procedures, \emph{i.e.}, multi-head self-attention (MHSA) module and feed-forward network (FFN). For the $i$-th block, the interaction process is:
\begin{equation}\label{eq2}
\begin{split}
    &{T_{rvl}^{(i-0.5)}}={T_{rvl}^{(i-1)}}+\mathrm{MHSA}({T_{rvl}^{(i-1)}}), \\
    &{T_{rvl}^{i}}={T_{rvl}^{(i-0.5)}}+\mathrm{FFN}({T_{rvl}^{(i-0.5)}}),
\end{split}
\end{equation}
where $T_{rvl}^{(0)}=T_{rvl}$. The attention operation for every head in $\mathrm{MHSA(\cdot)}$ is designed as:
\begin{equation}\label{sim}
\begin{split}
    {Y} & =\mathrm{Softmax}(\frac{QK^\mathsf{T}}{\sqrt{D}})V,\\
        {Q}  & = \phi_Q({T_{rvl}^{(i-1)}}),
    {K}  = \phi_K({T_{rvl}^{(i-1)}}),
    {V}  = \phi_V({T_{rvl}^{(i-1)}}),
\end{split}
\end{equation}
where $\phi_Q(\cdot)$, $\phi_K(\cdot)$, and $\phi_V(\cdot)$ are linear layers. 

For the Transformer encoder layer with FS, the computation process is formulated as:
\begin{equation}\label{eq2}
\begin{split}
    &{T_{rvl}^{(i-0.5)}}={T_{rvl}^{(i-1)}}+\mathrm{MHSA}({T_{rvl}^{(i-1)}}), \\
    &\hat{T}_{rvl}^{(i-0.5)} = \mathrm{FS}({T_{rvl}^{(i-0.5)}}), \\
    &{T_{rvl}^{i}}=\hat{T}_{rvl}^{(i-0.5)}+\mathrm{FFN}(\hat{T}_{rvl}^{(i-0.5)}),
\end{split}
\end{equation}
{where $i=4,7,10$ for CLIP-ViT-B consisting of 12 Transformer blocks} and $\mathrm{FS}(\cdot)$ is the feature selection procedure, which will be described in the next section.

As seen, for the proposed parallel structure, it has two key characteristics: 1) With the encapsulated input mechanism, the linguistic features are propagated through the entire visual feature extraction process from beginning to end, which allows the model to selectively focus on regions related to natural language expressions. This makes it possible and rational to execute the feature selection process later. 2) Compared to the existing pipeline with two stages, \emph{i.e.}, visual feature extraction and cross-modal information fusion, our proposed method is simpler. Attributed to the attention process on the encapsulated sequence, it is natural to achieve the sufficient interaction between text features and vision features, which makes it unnecessary to design additional feature fusion modules.

\subsection{Language-guided Visual Feature Selection}

Although the proposed parallel information interaction manner has the potential to extract effective and useful visual features for multimodal reasoning, 
the concatenation of visual tokens and linguistic tokens increases the length of the token sequence, which increases the computational complexity of the model. 
Inspired by DynamicViT~\cite{dynamicvit}, in this section, we introduce a language-guided visual feature selection mechanism $\mathrm{FS}(\cdot)$, which gradually discards visual tokens with low information density and low correlation with natural language representation during the feature extraction process.
In this manner, without affecting the visual feature extraction, the length of the token sequence would be shortened, reducing computational complexity and accelerating model inference.

\begin{figure}[t]
    \centering
    \includegraphics[width=0.8\linewidth]{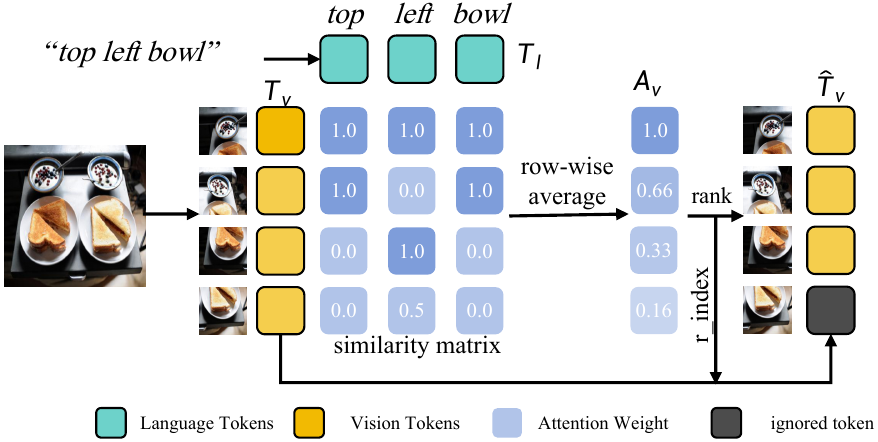}
    \caption{Diagram of Language-guided Visual Feature Selection.}
    \label{fig: select}
     \vspace{-4mm}
\end{figure}

As shown in Fig.~\ref{fig: select}, to select the linguistic-relevant visual tokens, one simple and intuitive selection strategy is based on the similarity matrix ${A \in \mathbb{R}^{(1+ N_v + N_l) \times (1+ N_v + N_l)}}$ between visual tokens and linguistic tokens, as:
\begin{equation}
\begin{split}
    A = QK^\mathsf{T} = \phi_Q({T_{rvl}^{(i-1)}})(\phi_K({T_{rvl}^{(i-1)}}))^\mathsf{T}.
\end{split}
\end{equation}
Based on the similarity matrix $A$, we can easily obtain the similarity $A_v\in \mathbb{R}^{N_{v}}$ between each visual token and all language tokens by averaging the similarity matrix on the dimension of linguistic tokens.
The higher the attention value, the more consistent the visual tokens are with the semantics of natural language expression. By ranking the attention value for every visual token, we can keep a certain percentage of visual tokens $\rho$ only. For the Transformer block in Eq.~\eqref{eq2}, the concrete computation with feature selection is: 
\begin{equation}\label{eq:fs}
\begin{split}
{T_{rvl}^{(i-0.5)}}&={T_{rvl}^{(i-1)}}+\mathrm{MHSA}({T_{rvl}^{(i-1)}}) \\
\mathrm{Split}({T_{rvl}^{(i-0.5)}})&\triangleq[T^{(i-0.5)}_{REG}, T^{(i-0.5)}_{v},T^{(i-0.5)}_{l}], \\
\mathrm{r\_index}&=\mathrm{rank}(A_{v}, \rho N_{v}), \\
\hat{T}^{(i-0.5)}_{v} &= [{T}^{(i-0.5)}_{v}]_{\mathrm{r\_index}}, \\
\hat{T}_{rvl}^{(i-0.5)}&=\mathrm{Concat}[T^{(i-0.5)}_{REG},\hat{T}^{(i-0.5)}_{v},T^{(i-0.5)}_{l}],\\
{T_{rvl}^{i}}~~~&=\hat{T}_{rvl}^{(i-0.5)}+\mathrm{FFN}(\hat{T}_{rvl}^{(i-0.5)}),
\end{split}
\end{equation}
where the second equation represents that the computed  $T_{rvl}^{(i-0.5)}$ can be partitioned into three parts along the channel dimension as $T^{(i-0.5)}_{REG}$, $T^{(i-0.5)}_{v}$, and $T^{(i-0.5)}_{l}$. $\mathrm{r\_index}$ is the row index set where the first $\rho N_{v}$ elements of $A_{v}$ are located, $[{T}^{(i-0.5)}_{v}]_{\mathrm{r\_index}}$ denotes extracting the corresponding submatrix from ${T}^{(i-0.5)}_{v} \in \mathbb{R}^{N_{v}\times D}$ according to the row index set ``$\mathrm{r\_index}$'', and $\hat{T}^{(i-0.5)}_{v} \in \mathbb{R}^{\rho N_{v}\times D}$.
The larger $\rho$ is, the more tokens are selected and the greater the computational overhead required. Please note that different from Dynamic ViT, instead of only adopting visual tokens, what we utilize is the similarity between language tokens and visual tokens for helping selecting useful visual tokens under the guidance of linguistic features.

\begin{figure*}[t]
    \centering
    \includegraphics[width=0.8\linewidth]{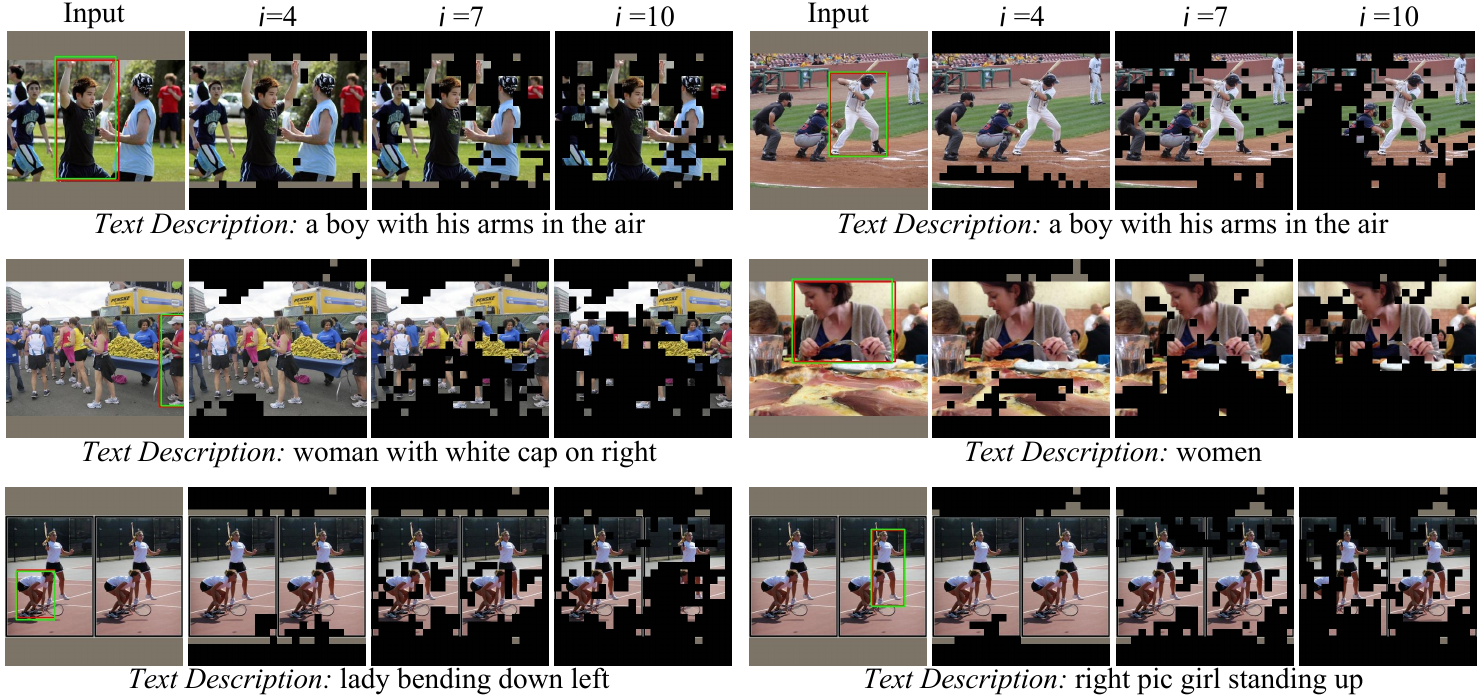}
     \vspace{2mm}
    \caption{{Visualization of language-guided visual feature selection on CLIP-ViT-B based on the RefCOCO val set}. For input image, the red bounding box is ground truth and the green box is the prediction of our proposed FSVG ($\rho=0.7$). The black patch is the discarded region which are decided by our proposed language-based visual feature selection process.}
    \label{fig: vts_results}
    \vspace{2mm}
\end{figure*}

To better understand our proposed $\mathrm{FS}(\cdot)$, based on the base version of ViT and the benchmark ReFCOCO val set, Fig.~\ref{fig: vts_results} visualizes the visual feature selection procedure for different Transformer blocks $i=4,7,10$.
As shown, our method can gradually understand the semantics of natural language expressions well, retain semantically consistent visual information, and then discard unimportant visual content. 
Besides, the third row of Fig.~\ref{fig: vts_results} (the same image, but different natural language expressions), shows that our method can accurately identify different positions and postures of similar targets.

\vspace{-1mm}
\subsection{Head and Training Loss}
\vspace{-1mm}
For localization, as shown in Fig.~\ref{fig: fsvg}, we use the [REG] token output of the vision backbone as the input of the head to predict the bounding box. For the head,  it consists of 3 linear layers with the input dimension as 256 and the output dimension as 4, which represent the center coordinates, width, and height of the predicted bounding box $\hat{b}=(\hat{c}_{x},\hat{c}_{y}, \hat{w},\hat{h})$.

Given the prediction $\hat{b}$ and the ground truth $b$, we adopt the following loss form to train FSVG as:
\vspace{-2mm}
\begin{equation}\label{eq3}
\begin{split}
    \mathcal{L} = \lambda_1 \mathcal{L}_{1}(b, \hat{b}) + \lambda_2 \mathcal{L}_{giou}(b, \hat{b}),
\end{split}
\end{equation}
where $\mathcal{L}_{1}(\cdot)$ and $\mathcal{L}_{giou}(\cdot)$ represent the $l_1$ loss and the generalized IoU loss, respectively, and $\lambda_1$ and $\lambda_2$ are the weighting coefficients for balancing different loss terms. In experiments, we empirically set $\lambda_1=5$ and $\lambda_2 = 2$.

\vspace{-2mm}
\section{Experiments}
\vspace{-2mm}
In this section, we evaluate our proposed FSVG based on a series of comparison experiments and ablation studies.

\begin{table*}[!ht]
\centering
\caption{Quantitative comparison with state-of-the-art methods on RefCOCO, RefCOCO+, RefCOCOg, and ReferIt. We highlight the best two results on each dataset in bold and underlined, respectively.}
\label{tab: sota}
\begin{tabular}{c|c|ccc|ccc|cc|c|c}
\hline
\multirow{2}{*}{Method} & \multirow{2}{*}{Venue} & \multicolumn{3}{c|}{RefCOCO} & \multicolumn{3}{c|}{RefCOCO+} & \multicolumn{2}{c|}{RefCOCOg} & ReferIt & \multirow{2}{*}{Avg} \\
                        &                   & val     & testA   & testB   & val      & testA   & testB   & val           & test                & test    \\
\hline
SAFF~\cite{saff}           & MM'21             & 79.26   & 81.09   & 76.55   & 64.43    & 68.46   & 58.43   & 68.94         & 68.91       & 66.01 & 70.23 \\
LBYL-Net~\cite{lbyl}       & CVPR'21           & 79.67   & 82.91   & 74.15   & 68.64    & 73.38   & 59.49   & -             & -           & 67.47 & - \\
Ref-TR~\cite{reftr}                 & NeurIPS’21        & 82.23   & 85.59   & 76.57   & 71.58    & 75.96   & 62.16   & 68.41         & 69.40       & 71.42 & 73.70 \\
TransVG~\cite{transvg}                 & ICCV’21           & 81.02   & 82.72   & 78.35   & 64.82    & 70.70   & 56.94   & 68.67         & 67.73       & 70.73 & 71.30 \\
SeqTR~\cite{seqtr}                   & ECCV’22           & 81.23   & 85.00   & 76.08   & 68.82    & 75.37   & 58.78   & 71.35         & 71.58       & 69.66 & 73.10 \\
Word2Pix~\cite{word2pix}                & TNNLS’22          & 81.20   & 84.36   & 78.12   & 69.74    & 76.11   & 61.24   & 70.81         & 71.34       & -     & - \\
QRNet~\cite{qrnet}            & CVPR’22       & 84.01   & 85.85   &\textbf{ 82.34}   & \underline{72.94}    & 76.17   & 63.81   & 73.03         & 72.52       & \textbf{74.61} & \underline{76.14} \\
CLIP-VG~\cite{clipvg}         & TMM’23        & 84.29   & \textbf{87.76}   & 78.43   & 69.55    & 77.33   & 57.62   & 73.18         & 72.54       & -     & - \\
JMRI~\cite{jmri}              & TIM'23        & 82.97   & 87.30   & 74.62   & 71.17    & 79.82   & 57.01   & 71.96         & 72.04       & 68.23 & 73.90 \\
RealGIN~\cite{realgin}        & TNNLS'23      & 80.38   & 81.08   & 77.25   & 62.90    & 65.50   & 57.40    & 65.52         & 65.57      & -     & - \\
LADS~\cite{lads}              & AAAI'23       & 82.85   & 86.67   & 78.57   & 71.16    & 77.64   & 59.82   & 71.56         & 71.66       & -     & - \\
ScanFormer~\cite{scanformer}  & CVPR'24       & 83.40   & 85.86   & 79.81   & 72.96    & 77.57   & 62.50   & \textbf{74.10}         & \textbf{74.14}       & 68.85 & 75.47 \\
CREC~\cite{crec}              & CVPR'24       & 82.77   & 86.35   & 77.13   & 72.29    & 78.24   & 63.47   & \underline{73.33}         & \underline{74.11}       & -     & - \\
\hline
{FSVG ($\rho=1$)}   & Ours & \textbf{84.59}   & \underline{87.40}    & 80.06   & \textbf{74.27}    & \textbf{80.64}   & \textbf{64.01}   & 72.75 & 73.15       & \underline{71.93} & \textbf{76.51} \\
{FSVG ($\rho=0.7$)} & Ours & \underline{84.45}& 87.19 & \underline{80.30} & 72.88 & \underline{79.93} & \underline{63.95} & 71.88 & 72.16 & 72.29 & 76.11 \\

\hline
\end{tabular}
\end{table*}

\begin{table}[h]
    \caption{Comparison on the number of model parameters and frames per second (FPS) of different methods with released source codes. Here FPS is averagely computed on RefCOCO val set {{with the image size as 384 $\times$ 384 based on an NVIDIA A6000 GPU.}}}
    \label{tab: fps}
    \centering
    \begin{tabular}{c|c|c|c}
    \hline
    {Method}  & {$\#$ Parameters}   & {FPS} & {Acc@0.5}  \\
        \hline
            TransVG    & \underline{170M}  & 15.00    & 81.02 \\
            QRNet      & 273M  & 10.97    & 84.01 \\
            CLIP-VG    & 181M  & 27.64    & 84.29 \\
            JMRI       & 216M  & -        & 82.97 \\
            ScanFormer & -  & 27$\sim$28    & 83.40 \\
            \hline
            FSVG ($\rho=1$)    & \textbf{150M} & \underline{29.85}    & \textbf{84.59} \\
            FSVG ($\rho=0.7$)  & \textbf{150M} & \textbf{30.52}    & \underline{84.45} \\
        \hline
    \end{tabular}
\end{table}

\begin{table}[!h]
    \caption{Effect of the ratio $\rho$ of feature selection on the performance on RefCOCO.}
    \label{tab: keeprate}
    \centering
    \begin{tabular}{c|c|ccc}
    \hline
\multirow{2}{*}{\makecell{Selection \\ Ratio}} & \multirow{2}{*}{GFLOPs} & \multicolumn{3}{c}{RefCOCO} \\
    & & val & testA & testB    \\
    \hline
    $\rho=1.0$ & 157.2G & 84.59 & 87.40 & 80.06  \\
    $\rho=0.9$ & 139.2G & 84.64 & 87.15 & 79.55  \\
    $\rho=0.8$ & 123.1G & 84.67 & 87.01 & 79.40  \\
    $\rho=0.7$ & 109.5G & 84.45 & 87.19 & 80.30  \\
    $\rho=0.6$ & 97.8G  & 83.97 & 87.45 & 79.35  \\
    $\rho=0.5$ & 87.9G  & 83.43 & 86.18 & 77.63  \\
    \hline
\end{tabular}
\end{table}



\subsection{Datasets and Evaluation Metric}

To comprehensively evaluate our approach, {four widely-adopted benchmark datasets} for visual grounding are used, including RefCOCO~\cite{refcoco}, RefCOCO+~\cite{refcoco}, RefCOCOg~\cite{refcocog}, and ReferIt~\cite{referit}. 
More information about the datasets can be found in the appendix.
Following~\cite{transvg++,yu2024revisiting,vg-law,chen2023advancing}, we use $Acc@0.5$ for quantitative evaluation. If the intersection-over-union (IoU) between the bounding box predicted by the model and the ground truth is greater than 0.5, we consider the predicted bounding box of the model to be correct.
\vspace{-1mm}
\subsection{Implementation Details}
\vspace{-1mm}
For FSVG,  we adopt the visual encoder and text encoder of CLIP~\cite{clip} for visual feature learning and linguistic feature learning, respectively.
Similar to JMRI~\cite{jmri}, CLIP-VG~\cite{RvG-Tree}, and ScanFormer~\cite{scanformer}, we choose the CLIP-ViT-B version.
Our experiments are implemented based on PyTorch by using two NVIDIA A100 GPUs. 
The model is end-to-end optimized by AdamW~\cite{adamw} and the weight decay is $1\times10^{-4}$. 
The number of the total training epochs set to 90 and the batch size is 128. 
For the visual backbone and language backbone, the initial learning rate is $1\times10^{-5}$. {For the head, the initial learning rate is $1\times10^{-4}$ and it decays by multiplying 0.1 at 60-th epoch}. 
The input image is resized to $384 \times 384$ pixels and the referring expressions are padded or truncated to 77 tokens. 
The ratio $\rho$ for visual feature selection in Eq.~\eqref{eq:fs} is set to 0.7. 
\subsection{Experimental Comparison}
Table~\ref{tab: sota} reports the quantitative results of different comparing methods on four benchmark datasets. In the case without feature selection as $\rho=1$, our proposed FSVG almost outperforms other baselines and achieves the higher average localization accuracy. When $\rho=0.7$, although only adopting partial visual features for faster computation speed, our proposed FSVG still achieves quite competitive performance across all the datasets, which finely substantiates the effectiveness of our proposed parallel structure as well as the  feature selection mechanism.

Table~\ref{tab: fps} compares the number of network parameters and the frames per second (FPS) of different methods with released source codes, including TransVG, QRNet, CLIP-VG, and our proposed FSVG.
Here the FPS is averagely computed based on an A6000 GPU on RefCOCO val set. 
Besides, for a full comparison, although ScanFormer~\cite{scanformer} has not released the code, based on the frames per second (FPS) provided in published papers, we find that its FPS is about 2.5 times that of QRNet, so we can roughly get the FPS of ScanFormer under our test configuration. 
It is easily observed that the proposed FSVG consistently outperforms these comparing methods, with faster inference speed, higher prediction accuracy, and fewer model parameters.
It is worth mentioning that although our FSVG is slightly inferior (QRNet and CLIP-VG) in some datasets as reported in Table~\ref{tab: sota}, it has higher computational efficiency, which is quite meaningful for practical applications.


\subsection{Ablation Study}

Table~\ref{tab: keeprate} reports the accuracy of our proposed FSVG on the four benchmark datasets under different values of $\rho$ for the visual feature selection in Eq.~\eqref{eq:fs}. We can find that 
as the ratio $\rho$ gets smaller, the number of selected visual features becomes smaller, thereby having lower GFLOPs. However, there is a general downward trend in performance. Especially, when $\rho$ is too small, like 0.6 and 0.5, the accuracy drops drastically. The underlying reason is that an extremely small $\rho$ would lead to the serious loss of target information. Considering the overhead and performance, we set $\rho$ as 0.7 in the experiments.

\textit{More details and comparison experiments as well as the related work are provided in the supplementary material.}
\section{Conclusion}
In this paper, we proposed a simple and effective feature selection-based visual grounding framework, called FSVG. The key specificity lies in: 1) We proposed to adopt the parallel structure to deal with the multi-modal features, which enables the full propagation of linguistic features to guide the important visual feature extraction, and avoids the additional cross-modal fusion module; 2) We constructed a feature selection mechanism to only utilize the linguistic-relevant visual features for prediction that makes it possible to obviously speed up the model computation process. Based on four benchmark datasets, extensive experiments substantiated the superiority of our FSVG in balancing accuracy and efficiency. 


\bibliographystyle{IEEEbib}
\bibliography{icme2025}

\begin{thebibliography}{10}

\bibitem{scanformer}
Wei Su, Peihan Miao, Huanzhang Dou, and Xi~Li,
\newblock ``Scanformer: Referring expression comprehension by iteratively scanning,''
\newblock in {\em Proceedings of the IEEE/CVF Conference on Computer Vision and Pattern Recognition}, 2024, pp. 13449--13458.

\bibitem{cyco}
Ning Wang, Jiajun Deng, and Mingbo Jia,
\newblock ``Cycle-consistency learning for captioning and grounding,''
\newblock in {\em Proceedings of the AAAI Conference on Artificial Intelligence}, 2024, vol.~38, pp. 5535--5543.

\bibitem{selfeq}
Ruozhen He, Paola Cascante-Bonilla, Ziyan Yang, Alexander~C Berg, and Vicente Ordonez,
\newblock ``Improved visual grounding through self-consistent explanations,''
\newblock in {\em Proceedings of the IEEE/CVF Conference on Computer Vision and Pattern Recognition}, 2024, pp. 13095--13105.

\bibitem{pvd}
Zesen Cheng, Kehan Li, Peng Jin, Siheng Li, Xiangyang Ji, Li~Yuan, Chang Liu, and Jie Chen,
\newblock ``Parallel vertex diffusion for unified visual grounding,''
\newblock in {\em Proceedings of the AAAI Conference on Artificial Intelligence}, 2024, vol.~38, pp. 1326--1334.

\bibitem{zeroshot_rec}
Zeyu Han, Fangrui Zhu, Qianru Lao, and Huaizu Jiang,
\newblock ``Zero-shot referring expression comprehension via structural similarity between images and captions,''
\newblock in {\em Proceedings of the IEEE/CVF Conference on Computer Vision and Pattern Recognition}, 2024, pp. 14364--14374.

\bibitem{transformer}
Ashish Vaswani, Noam Shazeer, Niki Parmar, Jakob Uszkoreit, Llion Jones, Aidan~N Gomez, {\L}ukasz Kaiser, and Illia Polosukhin,
\newblock ``Attention is all you need,''
\newblock {\em Advances in neural information processing systems}, vol. 30, 2017.

\bibitem{transvg++}
Jiajun Deng, Zhengyuan Yang, Daqing Liu, Tianlang Chen, Wengang Zhou, Yanyong Zhang, Houqiang Li, and Wanli Ouyang,
\newblock ``Transvg++: End-to-end visual grounding with language conditioned vision transformer,''
\newblock {\em IEEE transactions on pattern analysis and machine intelligence}, 2023.

\bibitem{qrnet}
Jiabo Ye, Junfeng Tian, Ming Yan, Xiaoshan Yang, Xuwu Wang, Ji~Zhang, Liang He, and Xin Lin,
\newblock ``Shifting more attention to visual backbone: Query-modulated refinement networks for end-to-end visual grounding,''
\newblock in {\em Proceedings of the IEEE/CVF Conference on Computer Vision and Pattern Recognition}, 2022, pp. 15502--15512.

\bibitem{lgrnet}
Mingcong Lu, Ruifan Li, Fangxiang Feng, Zhanyu Ma, and Xiaojie Wang,
\newblock ``Lgr-net: Language guided reasoning network for referring expression comprehension,''
\newblock {\em IEEE Transactions on Circuits and Systems for Video Technology}, 2024.

\bibitem{dynamicvit}
Yongming Rao, Wenliang Zhao, Benlin Liu, Jiwen Lu, Jie Zhou, and Cho-Jui Hsieh,
\newblock ``Dynamicvit: Efficient vision transformers with dynamic token sparsification,''
\newblock {\em Advances in neural information processing systems}, vol. 34, pp. 13937--13949, 2021.

\bibitem{saff}
Jiabo Ye, Xin Lin, Liang He, Dingbang Li, and Qin Chen,
\newblock ``One-stage visual grounding via semantic-aware feature filter,''
\newblock in {\em Proceedings of the 29th ACM International Conference on Multimedia}, 2021, pp. 1702--1711.

\bibitem{lbyl}
Binbin Huang, Dongze Lian, Weixin Luo, and Shenghua Gao,
\newblock ``Look before you leap: Learning landmark features for one-stage visual grounding,''
\newblock in {\em Proceedings of the IEEE/CVF conference on computer vision and pattern recognition}, 2021, pp. 16888--16897.

\bibitem{reftr}
Muchen Li and Leonid Sigal,
\newblock ``Referring transformer: A one-step approach to multi-task visual grounding,''
\newblock {\em Advances in neural information processing systems}, vol. 34, pp. 19652--19664, 2021.

\bibitem{transvg}
Jiajun Deng, Zhengyuan Yang, Tianlang Chen, Wengang Zhou, and Houqiang Li,
\newblock ``Transvg: End-to-end visual grounding with transformers,''
\newblock in {\em Proceedings of the IEEE/CVF International Conference on Computer Vision}, 2021, pp. 1769--1779.

\bibitem{seqtr}
Chaoyang Zhu, Yiyi Zhou, Yunhang Shen, Gen Luo, Xingjia Pan, Mingbao Lin, Chao Chen, Liujuan Cao, Xiaoshuai Sun, and Rongrong Ji,
\newblock ``Seqtr: A simple yet universal network for visual grounding,''
\newblock in {\em European Conference on Computer Vision}. Springer, 2022, pp. 598--615.

\bibitem{word2pix}
Heng Zhao, Joey~Tianyi Zhou, and Yew-Soon Ong,
\newblock ``Word2pix: Word to pixel cross-attention transformer in visual grounding,''
\newblock {\em IEEE Transactions on Neural Networks and Learning Systems}, vol. 35, no. 2, pp. 1523--1533, 2022.

\bibitem{clipvg}
Linhui Xiao, Xiaoshan Yang, Fang Peng, Ming Yan, Yaowei Wang, and Changsheng Xu,
\newblock ``Clip-vg: Self-paced curriculum adapting of clip for visual grounding,''
\newblock {\em IEEE Transactions on Multimedia}, 2023.

\bibitem{jmri}
Hong Zhu, Qingyang Lu, Lei Xue, Mogen Xue, Guanglin Yuan, and Bineng Zhong,
\newblock ``Visual grounding with joint multi-modal representation and interaction,''
\newblock {\em IEEE Transactions on Instrumentation and Measurement}, 2023.

\bibitem{realgin}
Yiyi Zhou, Rongrong Ji, Gen Luo, Xiaoshuai Sun, Jinsong Su, Xinghao Ding, Chia-Wen Lin, and Qi~Tian,
\newblock ``A real-time global inference network for one-stage referring expression comprehension,''
\newblock {\em IEEE Transactions on Neural Networks and Learning Systems}, vol. 34, no. 1, pp. 134--143, 2021.

\bibitem{lads}
Wei Su, Peihan Miao, Huanzhang Dou, Yongjian Fu, and Xi~Li,
\newblock ``Referring expression comprehension using language adaptive inference,''
\newblock in {\em Proceedings of the AAAI Conference on Artificial Intelligence}, 2023, vol.~37, pp. 2357--2365.

\bibitem{crec}
Zhihan Yu and Ruifan Li,
\newblock ``Revisiting counterfactual problems in referring expression comprehension,''
\newblock in {\em Proceedings of the IEEE/CVF Conference on Computer Vision and Pattern Recognition}, 2024, pp. 13438--13448.

\bibitem{refcoco}
Licheng Yu, Patrick Poirson, Shan Yang, Alexander~C Berg, and Tamara~L Berg,
\newblock ``Modeling context in referring expressions,''
\newblock in {\em Computer Vision--ECCV 2016: 14th European Conference, Amsterdam, The Netherlands, October 11-14, 2016, Proceedings, Part II 14}. Springer, 2016, pp. 69--85.

\bibitem{refcocog}
Varun~K Nagaraja, Vlad~I Morariu, and Larry~S Davis,
\newblock ``Modeling context between objects for referring expression understanding,''
\newblock in {\em Computer Vision--ECCV 2016: 14th European Conference, Amsterdam, The Netherlands, October 11--14, 2016, Proceedings, Part IV 14}. Springer, 2016, pp. 792--807.

\bibitem{referit}
Sahar Kazemzadeh, Vicente Ordonez, Mark Matten, and Tamara Berg,
\newblock ``Referitgame: Referring to objects in photographs of natural scenes,''
\newblock in {\em Proceedings of the 2014 conference on empirical methods in natural language processing (EMNLP)}, 2014, pp. 787--798.

\bibitem{yu2024revisiting}
Zhihan Yu and Ruifan Li,
\newblock ``Revisiting counterfactual problems in referring expression comprehension,''
\newblock in {\em Proceedings of the IEEE/CVF Conference on Computer Vision and Pattern Recognition}, 2024, pp. 13438--13448.

\bibitem{vg-law}
Wei Su, Peihan Miao, Huanzhang Dou, Gaoang Wang, Liang Qiao, Zheyang Li, and Xi~Li,
\newblock ``Language adaptive weight generation for multi-task visual grounding,''
\newblock in {\em Proceedings of the IEEE/CVF conference on computer vision and pattern recognition}, 2023, pp. 10857--10866.

\bibitem{chen2023advancing}
Zhihong Chen, Ruifei Zhang, Yibing Song, Xiang Wan, and Guanbin Li,
\newblock ``Advancing visual grounding with scene knowledge: Benchmark and method,''
\newblock in {\em Proceedings of the IEEE/CVF Conference on Computer Vision and Pattern Recognition}, 2023, pp. 15039--15049.

\bibitem{clip}
Alec Radford, Jong~Wook Kim, Chris Hallacy, Aditya Ramesh, Gabriel Goh, Sandhini Agarwal, Girish Sastry, Amanda Askell, Pamela Mishkin, Jack Clark, et~al.,
\newblock ``Learning transferable visual models from natural language supervision,''
\newblock in {\em International conference on machine learning}. PMLR, 2021, pp. 8748--8763.

\bibitem{RvG-Tree}
Richang Hong, Daqing Liu, Xiaoyu Mo, Xiangnan He, and Hanwang Zhang,
\newblock ``Learning to compose and reason with language tree structures for visual grounding,''
\newblock {\em IEEE transactions on pattern analysis and machine intelligence}, vol. 44, no. 2, pp. 684--696, 2019.

\bibitem{adamw}
Ilya Loshchilov and Frank Hutter,
\newblock ``Decoupled weight decay regularization,''
\newblock in {\em International Conference on Learning Representations}, 2019.

\end{thebibliography}
\end{document}